\documentclass{article}
\usepackage{spconf,amsmath,epsfig}

\usepackage{amssymb}
\usepackage{hyperref}
\usepackage{color}
\usepackage{booktabs}
\usepackage{graphicx}
\graphicspath{{./figures/}}
\DeclareGraphicsExtensions{.eps,.pdf,.jpeg,.png,.tif}

\usepackage[caption=false,font=normalsize]{subfig}

\newcommand\Mark[1]{\textsuperscript#1}

\title{Building Footprint Extraction with Graph Convolutional Network}
\name{Yilei Shi \Mark{1}, Qinyu Li \Mark{2}, Xiaoxiang Zhu \Mark{2}\Mark{,}\Mark{3}}
\address{\Mark{1} Chair of Remote Sensing Technology (LMF), Technical University of Munich, Munich, Germany \\
\Mark{2} Signal Processing in Earth Observation (SiPEO), Technical University of Munich, Munich, Germany \\
\Mark{3} Remote Sensing Technology Institute (IMF), German Aerospace Center (DLR), Wessling, Germany}

\begin{document}
%
\maketitle
\begin{abstract}
\textcolor{blue}{This is the pre-acceptance version, to read the final version please go to IEEE XPlore.} Building footprint information is an essential ingredient for 3-D reconstruction of urban models. The automatic generation of building footprints from satellite images presents a considerable challenge due to the complexity of building shapes. Recent developments in deep convolutional neural networks (DCNNs) have enabled accurate pixel-level labeling tasks. One central issue remains, which is the precise delineation of boundaries. Deep architectures generally fail to produce fine-grained segmentation with accurate boundaries due to progressive downsampling. In this work, we have proposed a end-to-end framework to overcome this issue, which uses the graph convolutional network (GCN) for building footprint extraction task. Our proposed framework outperforms state-of-the-art methods.
\end{abstract}
\section{Introduction}

Building footprint generation is of great importance to urban planning and monitoring, land use analysis, and disaster management. High-resolution satellite imagery, which can provide more abundant detailed ground information, has become a major data source for building footprint generation. Due to the variety and complexity of buildings, building footprint requires significant time and high costs to generate manually. As a result, the automatic generation of a building footprint not only minimizes the human role in producing large-scale maps but also greatly reduces time and costs.


Over the past few years, the most popular and efficient classification approach has been deep learning (DL) \cite{bib:zhu2017deep}, which has the computational capability for big data. DL methods combine feature extraction and classification and are based on the use of multiple processing layers to learn good feature representation automatically from the input data. Therefore, DL usually possesses better generalization capability, compared to other classification-based methods. In terms of particular DL architectures, several impressive convolutional neural network (CNN) structures, such as ResNet \cite{bib:he2016} and U-Net \cite{bib:ronneberger2015}, have already been widely explored for RS tasks.

Many deep learing methods have been developed for building footprint generation. In \cite{bib:yuan2018}, authors propose a multistage ConvNet with an upsampling operation of bilinear interpolation. The trained model achieves a superior performance on very-high-resolution aerial imagery. Recently, an end-to-end trainable active contour model (ACM) was developed for building instance extraction \cite{bib:marcos2018}, which learns ACM parameterizations using a DCNN. In \cite{bib:shi2018}, authors exploit the improved conditional Wasserstein generative adversarial network to generate the building footprint automatically. Recent work \cite{bib:wang2017} shows that most of the tasks, such as building segmentation, building height estimation, and building contour extraction, are still difficult for modern convolutional networks. In this work, we show a significant performance improvement of building footprint extraction by using our proposed novel framework.

\begin{figure*}
  \centering
  \includegraphics[width=\textwidth]{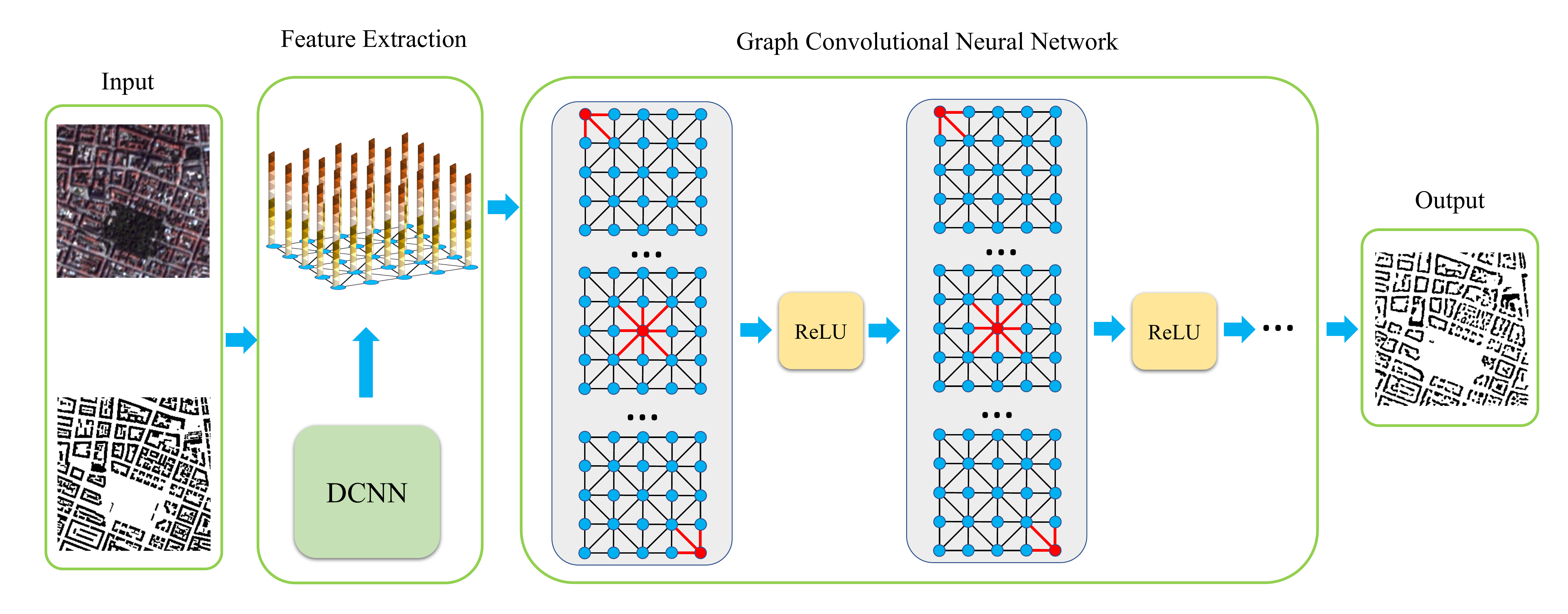}
  \caption{The framework of graph convolutional network}
  \label{fig:GCN_workflow}
\end{figure*}

\section{Methodology}
\subsection{Review of semantic segmentation}
Semantic segmentation with a fully convolutional network (FCN) was first introduced in \cite{bib:long2015},  which replaces the last few fully connected layers by convolutional layers to make efficient end-to-end learning and inference that can take arbitrary input size. In \cite{bib:badrinarayanan2015}, SegNet was proposed, which used an alternative decoder variant. The decoder uses pooling indices computed in the max-pooling step of the corresponding encoder to perform nonlinear upsampling. This makes SegNet more memory efficient than FCN.  Another variant of the encoder--decoder architecture is U-Net \cite{bib:ronneberger2015}. The architecture by its skip connections allows the decoder at each stage to learn back relevant features that are lost when pooled in the encoder.

One issue in FCN approaches is that by propagating through several alternated convolutional and pooling layers, the resolution of the output feature maps is downsampled. In order to overcome the poor localization property, \cite{bib:chen2017} offered an alternative to raise the output resolution, which used a probabilistic graph model CRF to refine the object boundary. CRFasRNN \cite{bib:zheng2015} extended to an end-to-end trainable network by introducing a fully connected CRF. In this work, we extended DCNNs to topologies that differ from the low-dimensional grid structure. The grid-like data can be viewed as a special type of graph data, where each node has a fixed number of ordered neighbors.

\subsection{Proposed method}
An undirected and connected graph $\mathcal{G} = (\mathcal{V}, \mathcal{E})$ consists of a set of nodes $\mathcal{V}$ and edges $\mathcal{E}$. The unnormalized graph Laplacian matrix $\mathbf{L}$ is defined as
\begin{equation}
 \mathbf{L} = \mathbf{D} - \mathbf{A},
\end{equation}
where $\mathbf{A}$ is the adjacency matrix representing the topology of $\mathcal{G}$ and $\mathbf{D}$ is the degree matrix with $D_{ii} = \sum_j A_{ij}$. As the graph Laplacian matrix $\mathbf{L}$ is a symmetric positive semi-definite matrix, its eigenvalue decomposition can be expressed as
\begin{equation}
  \mathbf{L} = \boldsymbol{\Phi} \boldsymbol{\Lambda} \boldsymbol{\Phi}^{T},
\end{equation}
where $\boldsymbol{\Phi} = \left( \phi_1, \phi_2, ..., \phi_n \right)$ are the orthonormal eigenvectors, known as the graph Fourier modes and $\boldsymbol{\Lambda} = \mathrm{diag} \left(\lambda_1, \lambda_2, ..., \lambda_n \right)$ is the diagonal matrix of corresponding non-negative eigenvalues. Assuming a signal $\mathbf{f}$ on the graph nodes $\mathcal{V}$, its graph Fourier transform is then defined as $\hat{\mathbf{f}} = \boldsymbol{\Phi}^T \mathbf{f}$. If $\mathbf{g}$ is a filter, the convolution of $\mathbf{f}$ and $\mathbf{g}$ can be written as
\begin{equation}
  \mathbf{f} * \mathbf{g} = \boldsymbol{\Phi} \left( \left( \boldsymbol{\Phi}^T \mathbf{g}\right) \circ \left( \boldsymbol{\Phi}^T \mathbf{f}\right) \right) = \boldsymbol{\Phi} \hat{\mathbf{g}} \boldsymbol{\Phi}^T \mathbf{f},
\end{equation}
where $\hat{\mathbf{g}}$ is the spectral representation of the filter. Rather than computing the Fourier transform $\hat{\mathbf{g}}$, the filter coefficients can be parameterized as $\hat{\mathbf{g}} = \sum \limits_{k=0}^r \alpha_k \beta_k$ in \cite{bib:henaff2015}. With the polynomial parametrization of the filter, the spectral filter is exactly localized in space, and its learning complexity is same as classical DCNNs. However, even with such a parameterization of the filters, the spectral GCN still suffers a high computational complexity.

Instead of explicitly operating in the frequency domain with a spectral multiplier, it is possible to represent the filters via a polynomial expansion $\hat{\mathbf{g}} = g(\boldsymbol{\Lambda})$ with the Chebyshev basis.
\begin{equation}
  g(\boldsymbol{\Lambda}) = \sum \limits_{k=0}^r \alpha_k T_k(\tilde{\boldsymbol{\Lambda}}),
\end{equation}
where $T_k(\tilde{\boldsymbol{\Lambda}})$ is the Chebyshev polynomials. The convolution can be formulated as
\begin{equation}
  \mathbf{f} * \mathbf{g} = \sum \limits_{k=0}^r \alpha_k T_k(\tilde{\mathbf{L}})\mathbf{f},
\end{equation}
where $\tilde{\mathbf{L}} = 2/\lambda_{max} \cdot \mathbf{L} - \mathbf{I}$ and $\lambda_{max}$ is the maximal eigenvector.
In \cite{bib:kipf2016}, the authors further simplify the Chebyshev framework, setting $r = 1$ and assuming $\lambda_{max} \approx 2$, allowing them to redefine a single convolutional layer.\\

\noindent \textbf{The propogation model}

\noindent The proposed propogation model can be written as
\begin{equation}
  \mathbf{H}_i^{l} = \sigma_r \left( \tilde{\mathbf{D}}^{-1/2} \tilde{\mathbf{A}} \tilde{\mathbf{D}}^{-1/2} \mathbf{W} \mathbf{H}_i^{l-1} \right)
\end{equation}
where $\mathbf{H}^{l}$ is the matrix of activations in the $l^{th}$ layer. $\tilde{\mathbf{A}} = \mathbf{A} + \mathbf{I}$ is the adjacency matrix of the undirected graph $\mathcal{G}$ with added self-connections. $\mathbf{I} $ is the identity matrix, $\tilde{D}_{ii} = \sum_j \tilde{A}_{ij}$, and $\mathbf{W}$ is the trainable weight matrix. $\sigma_r(\cdot)$ denotes a nonlinear activation function. This simplified form improves computational performance on larger graphs and predictive performance on small training sets.

\begin{equation}
  p = \mathrm{softmax}(\mathbf{H}_i^{l}) 
\end{equation}

\begin{figure*}
  \centering
  \subfloat[]{\includegraphics[width=0.16\textwidth]{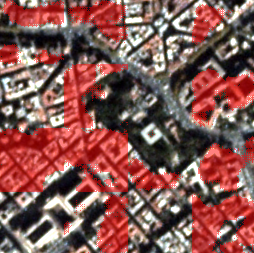}}
  \hfil
  \subfloat[]{\includegraphics[width=0.16\textwidth]{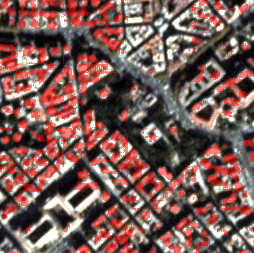}}
  \hfil
  \subfloat[]{\includegraphics[width=0.16\textwidth]{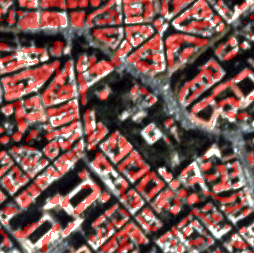}}
  \hfil
  \subfloat[]{\includegraphics[width=0.16\textwidth]{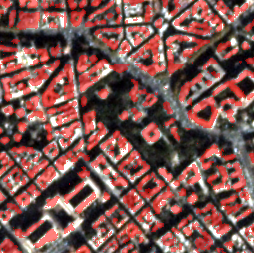}}
  \hfil
  \subfloat[]{\includegraphics[width=0.16\textwidth]{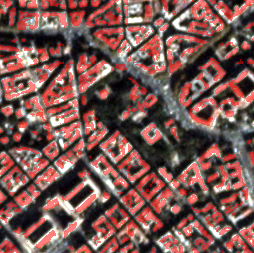}}
  \hfil
  \subfloat[]{\includegraphics[width=0.16\textwidth]{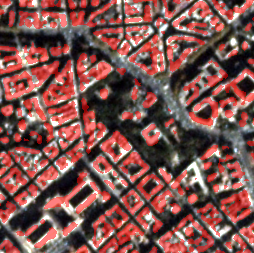}}
  \vfil
  \subfloat[]{\includegraphics[width=0.16\textwidth]{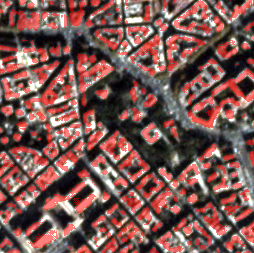}}
  \hfil
  \subfloat[]{\includegraphics[width=0.16\textwidth]{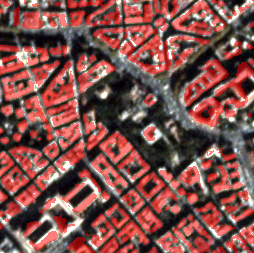}}
  \hfil
  \subfloat[]{\includegraphics[width=0.16\textwidth]{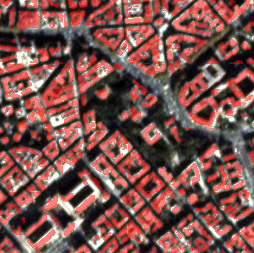}}
  \hfil
  \subfloat[]{\includegraphics[width=0.16\textwidth]{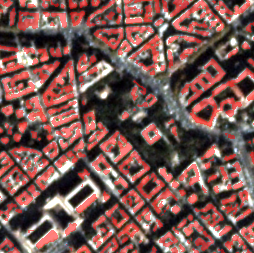}}
  \hfil
  \subfloat[]{\includegraphics[width=0.16\textwidth]{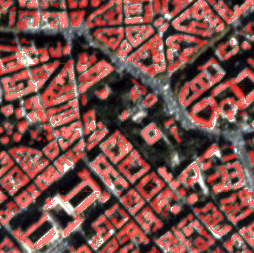}}
  \hfil
  \subfloat[]{\includegraphics[width=0.16\textwidth]{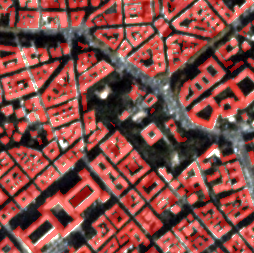}}
  \caption{Visualized comparison of the predicted results using different networks. (a) FCN-32s; (b) FCN-16s; (c) ResNet-DUC; (d) E-Net; (e) SegNet; (f) U-Net; (g) FCN-8s; (h) CWGAN-GP; (i) FC-DenseNet; (j) CRFasRNN; (k) GCN; (o) ground truth. }
  \label{fig:comp_different_methods}
\end{figure*}

\section{Experiments}

\subsection{Datasets}
In this work, we use Planetscope satellite images \cite{bib:planet} with RGB bands at a 3 m spatial resolution. The imagery is acquired by Doves, which form a satellite constellation that provides a complete image of the earth once per day. The study sites cover four cities: (1) Munich, Germany; (2) Rome, Italy; (3) Paris, France; (4) Zurich, Switzerland. The corresponding building footprint layer was downloaded from OpenStreetMap (OSM). The imagery is processed using a $\textrm{64} \times \textrm{64}$ sliding window with a stride of 19 pixels to produce 48,000 sample patches. The training data has 80\% patches, and the testing data has 20\% patches. The training and testing data are spatially separated.

\subsection{Experimental Setup}
For all networks, a stochastic gradient descent (SGD) with a learning rate of $\mathrm{10}^{-4}$ was adopted as an optimizer and negative log-likelihood loss (NLLLoss) was taken as the loss function. The implementation is based on the Pytorch and runs on a single NVIDIA Tesla P100 16 GB GPU. Semantic segmentation methods based on FCN-32s, FCN-16s, FCN-8s, ResNet-DUC, E-Net, SegNet, U-Net, CWGAN-GP, FC-DenseNet, GCN were taken as the algorithms of comparison.

\subsection{Results and Analysis}

In this work, we evaluated the inference performances using metrics for a quantitative comparison: overall accuracy (OA), F1 scores, and the Intersection over Union (IoU) scores. We evaluate the performance of different deep convolutinal neural networks and compare to our proposed method. The quantitative results are listed in Table \ref{tab:comp_dcnn_results}, and results of the sample for visual comparison are in Fig. \ref{fig:comp_different_methods}.

\begin{table}[h!]
\begin{center}
\begin{tabular}{cccc}
\toprule
\toprule
Methods  & \textbf{OA} & \textbf{F1} & \textbf{IoU} \\
\midrule
\rule{0pt}{2.5ex} FCN-32s  & 0.7318 & 0.2697 & 0.1559  \\
\rule{0pt}{2.5ex} FCN-16s  & 0.7698 & 0.3993 & 0.2494  \\
\rule{0pt}{2.5ex} ResNet-DUC  & 0.7945 & 0.4542 & 0.2930  \\
\rule{0pt}{2.5ex} E-Net  & 0.8243  & 0.5427 & 0.3724 \\
\rule{0pt}{2.5ex} SegNet & 0.8261  & 0.5558 & 0.3848 \\
\rule{0pt}{2.5ex} U-Net & 0.8412 & 0.6043  & 0.4329 \\
\rule{0pt}{2.5ex} FCN-8s & 0.8472 & 0.6222 & 0.4513 \\
\rule{0pt}{2.5ex} CWGAN-GP  & 0.8483 & 0.6268 & 0.4562 \\
\rule{0pt}{2.5ex} FC-DenseNet & 0.8551 & 0.6328 & 0.4628  \\

\rule{0pt}{2.5ex} CRFasRNN & 0.8592 & 0.6415  & 0.4757  \\

\rule{0pt}{2.5ex} GCN & \textbf{0.8640} & \textbf{0.6677} & \textbf{0.5012}  \\
\bottomrule
\bottomrule
\end{tabular}
\end{center}
\caption{Comparison of different deep convolutional neural networks on the test datasets}
\label{tab:comp_dcnn_results}
\end{table}

FCN-32s and FCN-16s exhibit poor performance, since the feature map of later layers have only high-level semantics with poor localization. ResNet-DUC can achieve better results than the previous two because of hybrid dilated convolution and dense upsampling convolution. It is limited due to the lack of skip connections. Max-pooling indices are reused in SegNet during the decoding process, which can reduce the number of parameters enabling end-to-end training. However, since it only uses max-pooling indices to decoder, some local details cannot be recovered, e.g. small buildings will be neglected. FCN-8s and U-Net outperform previous networks due to the concatenation of low-level features. Compared to other CNN models, CWGAN-GP shows promising results for building footprint generation. The skip connections in the generator combine both the lower and higher layers to generate the final output, retaining more details and better preserving the boundary of the building area. Moreover, the min-max game between the generator and discriminator of the GAN can motivate both to improve their functionalities. FC-DenseNet has better performance than previous networks, since DenseNet block concatenates feature maps learned by different layers, which can increase variation in the input of subsequent layers and improve efficiency.

GCN outperforms all other semantic segmentation neural networks in numerical accuracy and visual results. On one hand, it can aggregate the information from neighbor nodes (short range), which allows the model to learn about local structures. On the other hand, the DCNN can extract more comprehensive and representative features which enhance the feature fusion by embedding more spatial information into high-level features.

\section{Conclusion}
In this work, we develop a novel framework for semantic segmentation, which combines the DCNN and the GCN. Our proposed framework outperforms the state-of-the-art approaches for building footprint extraction. Furthermore, the proposed framework will be applied for the semantic segmentation of 3-D point clouds, which could be considered as a general graph.

\section{Acknowledgment}
This work is supported by the Bavaria California Technology Center (Project: Large-Scale Problems in Earth Observations, the European Research Council (No. ERC-2016-StG-714087, So2Sat), Helmholtz Association Young Investigators Group "SiPEO" (VHNG-1018, www.sipeo.bgu.tum.de). The authors thank the Gauss Centre for Supercomputing (GCS) and the Leibniz Supercomputing Centre (LRZ). The authors thank Planet provide the datasets.

\end{document}